\documentclass[10pt,a4paper]{article}
\usepackage[utf8]{inputenc}
\usepackage{amsmath}
\usepackage{amsfonts}
\usepackage{amssymb}
\usepackage{multirow}
\usepackage{graphicx}
\usepackage[left=2cm,right=2cm,top=2cm,bottom=2cm]{geometry}
\begin{document}
\title{Automatic Feature Weight Determination using Indexing and Pseudo Relevance Feedback for Multi-feature Content Based Image Retrieval}

\author{Asheet Kumar \and Shivam Choudhary \and Vaibhav Singh Khokhar \and Vikas Meena \and Chiranjoy Chattopadhyay\\
Indian Institute of Technology Jodhpur}

\maketitle

\begin{abstract}
Content-based image retrieval (CBIR) is one of the most active research area in multimedia information retrieval. Given a query image, the task is to search relevant images in a repository. Low level features like color, texture and shape feature vectors of an image are always considered to be an important attribute in CBIR system. Thus the performance of the CBIR system can be enhanced by combining these feature vectors. In this paper, we propose a novel CBIR framework by applying indexing using multiclass SVM and finding the appropriate weights of the individual features automatically using the relevance ratio and mean difference. We have taken four feature descriptors to represent color, texture and shape features. During retrieval, feature vectors of query image are combined, weighted and compared with feature vectors of images in the database to rank order the results. Experiments were performed on four benchmark datasets and performance is compared with existing techniques to validate the superiority of our proposed framework.
\end{abstract}

\section{Introduction}
\label{sec:intro}
CBIR has long been an active area of research within the multimedia information retrieval community.
Over the years, CBIR systems have evolved in complexity and performance. Among several open research issues in CBIR, use of multiple image features in conjunction is the most important one. Studies reveal that it is difficult to find a single feature which performs well across all possible scenarios, whereas using multiple features in conjunction has proven to be a good strategy. The main concern when using multiple features is how to weigh these features. One obvious method is giving equal weights. In tasks like image classification and object recognition, there is usually a training phase when feature weights can be ascertained. However, in image retrieval we have the practical problem of not knowing the relevance class of an image as not having a relevant training set. This limitation has forced the development of different relevance feedback techniques, which has its own very rich literature.
\subsection{Previous works on CBIR}
\label{litsurv}
In the literature there is a plethora of work which deals with CBIR. However, in this paper, we are highlighting the ones which are relevant to our work.

A probabilistic framework for efficient retrieval and indexing of image collections was proposed in \cite{hybrid}. This framework uncovers the hierarchical structure underlying the collection from image features based on a hybrid model that combines both generative and discriminative learning. The generalized Dirichlet mixture and maximum likelihood model was explored in \cite{hybrid} for the generative learning in order to estimate accurately the statistical model of the data. In \cite{mostrel}, a model for content-based image retrieval CBIR is proposed which depends only on extracting the most relevant features according to a feature selection technique. The suggested feature selection technique aims at selecting the optimal features that not only maximize the detection rate but also simplify the computation of the image retrieval process. In \cite{graph}, a novel CBIR scheme was proposed that exploits statistical features computed using the Multi-scale Geometric Analysis (MGA) of Non-subsampled Contourlet Transform (NSCT). 

A Comparative Study on feature extraction using texture and shape for CBIR is given in \cite{featurecomp}. The significance of the Local Binary Pattern (LBP) feature was discussed in \cite{LBP}. In \cite{cdh}, a novel image feature representation method using color difference histograms (CDH) was proposed for image retrieval. A novel CBIR approach was proposed in \cite{wavelet}, which uses a well-known clustering algorithm k-means and a database indexing structure B+ tree to facilitate efficient retrieval of relevant images. Cluster validity analysis indexes combined with majority voting are employed to verify the appropriate number of clusters. Minimum distance criteria was used to identify image cluster(s) closer to the query image. Daubechies wavelet transformation was used for extracting feature vectors from the images. The work proposed in \cite{multifeature} and \cite{featureweight} also gives weights to features and improves the accuracy of CBIR system. The work proposed in \cite{adapt}, presents a hybrid approach to reduce the semantic gap between low level visual features and high level semantics, through simultaneous feature adaptation and feature selection.A mixed gravitational search algorithm (MGSA) is proposed by the authors of \cite{adapt}. Feature extraction parameters are optimized to reach a maximum precision of the CBIR systems. An extremely fast CBIR system was proposed in \cite{multisvm}, which uses Multiple Support Vector Machines Ensemble. 

RF is an effective approach \cite{wang2014study} to bridge the gap between low-level visual features and high-level semantic meanings in CBIR. To reduce the computational complexity of traditional SVM based RF, an active SVM-based RF using multiple classifiers ensemble and features re-weighting is proposed in \cite{svmrf}. At first the most informative images are selected by using active learning method, then the feature space is modified dynamically by appropriately weighting the descriptive features according to a set of statistical characteristics. Finally the weight vectors of component SVM classifiers are computed dynamically by using the parameters for positive and negative samples. A comparative study of major challenges in RF for CBIR is given in \cite{rfcomp}. Authors of \cite{longrf} have proposed a long term learning scheme in relevance feedback for CBIR. 

\begin{figure}
\centering
\includegraphics[width=\linewidth]{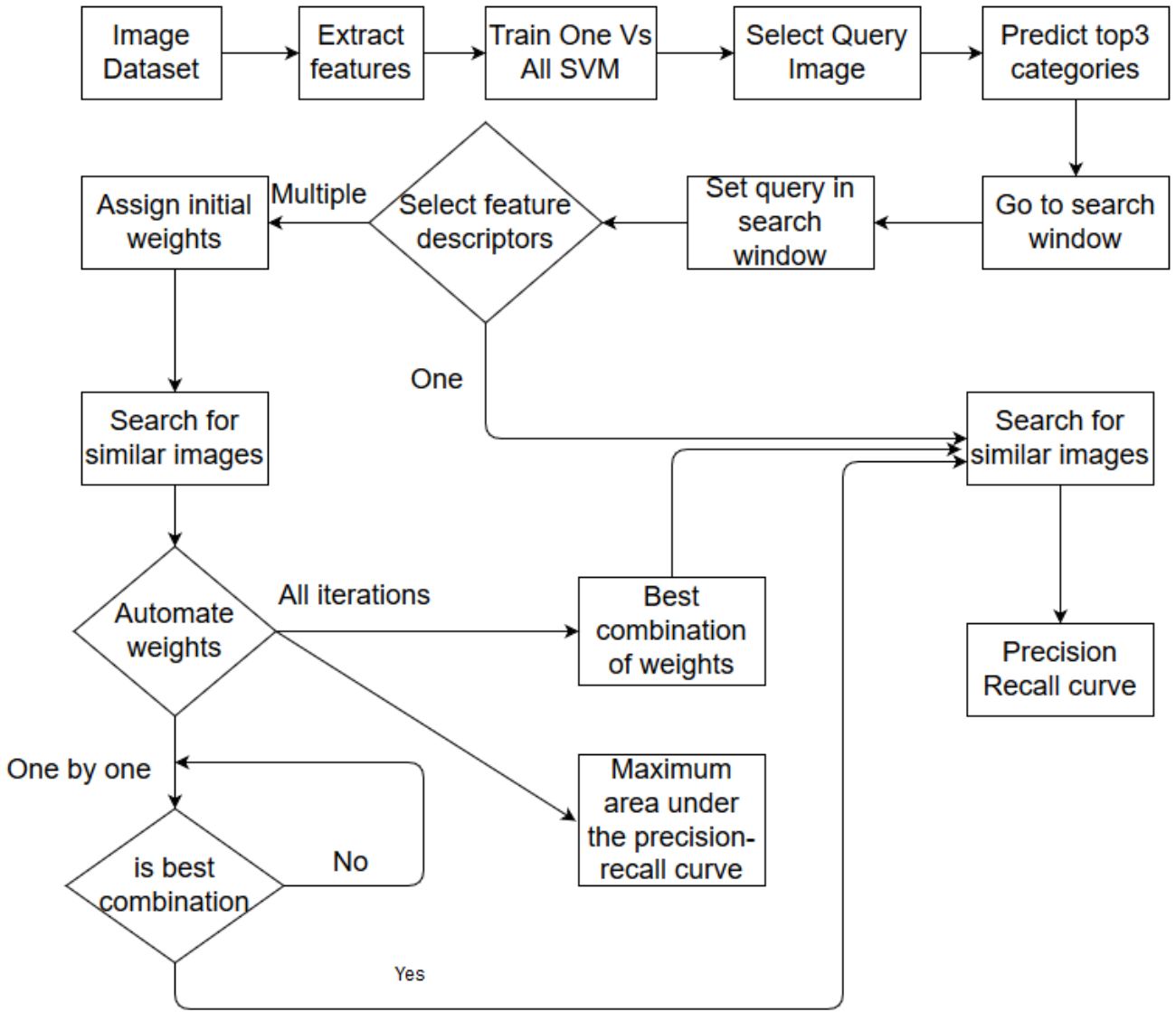}
\caption{Flow diagram of our proposed multi-feature CBIR framework.}
\label{fig1}     
\end{figure}

A novel perspective to retrieval partial-duplicate images with Contented-based Saliency Region (CSR) is proposed in \cite{saliency}. The content of CSR is represented with the BOV model while saliency analysis is employed to ensure the high visual attention of CSR. A relative saliency ordering constraint is also proposed by the authors that captures a weak saliency relative layout among interest points in the saliency regions.
\subsection{Contribution}
\label{intro:2}
The primary contribution of this paper are: (i) design of a multi-feature CBIR system, (ii) automatic weight selection of individual features, (iii) extensive experiments of four publicly available datasets and comparison with state-of-the-art techniques. 

\subsection{Organization of the paper}
\label{intro:3}
The organization of the paper is as follows: Section \ref{desc} briefly describes our proposed CBIR framework. In Sec. \ref{sec:feature}, we describe the four different feature descriptors used in this paper, along with the appropriate distance metric. Our proposed indexing technique is discussed in Sec. \ref{sec:index}. Next in Sec. \ref{sec:weight}, we describe the two different automatic feature weight selection methods. The experimental results are discussed in Sec. \ref{sec:result}. The positives and drawbacks of the proposed framework is given in Sec. \ref{sec:discuss}, followed by the concluding remarks in Sec. \ref{sec:concl}.
\section{Brief description of work}
\label{desc}

 Figure \ref{fig1} depicts the flow diagram of our proposed CBIR framework. We extract features of images in dataset and create a database of feature descriptors. Then we use features from database to train classifier model. We have used $60\%$ of features for training set and $40 \%$ for testing set. Once we have trained classifier, we select a query image and predict top $3$ categories (empirically determined). By doing this we have reduced search space for similar images. To search for similar images we select one or all feature descriptors. In case we selected all feature descriptors we assign them initial weights and then search for similar images using initial combination of weights. Then, we iteratively obtain the optimal weights for the individual feature descriptors. We retrieve similar images with optimally assigned weights and compute the PR curve to analyse the performance.

\section{Feature Descriptors}
\label{sec:feature}
A brief description of the feature descriptors used here, is given in the following sub-sections.
\subsection{Color Difference Histogram (CDH)} CDH counts the perceptually uniform color difference between two points under different backgrounds with regard to the colors and edge orientations in L*a*b* color space. 
Implementing CDH includes conversion from RGB to CLE L*a*b* colour space, detection of edge orientation, colour quantization in CLE L*a*b* colour space. This method pays more attention to color, edge orientation and perceptually uniform color differences, and encodes color, orientation and perceptually uniform color difference via feature representation in a similar manner to the human visual system. 
\subsection{Local Binary Pattern (LBP) } The generic local binary pattern operator is derived from joint distribution. As in the case of basic LBP is obtained by summing the thresholded differences weighted by powers of two. The $LBP_{P,R}$ operator is defined as
\begin{equation}
\textit(LBP)_{P,R}({x}_{c},{y}_{c})=\sum_{p=0}^{p-1} {s(g_{p}-g_{c})2^p}
\end{equation}
In this equation the signs of the differences in a neighbourhood are interpreted as a P-bit binary number, resulting in $2^{P}$ distinct values for the LBP code as:
\begin{equation}
\textit{T}\approx t(LBP_{P,R}({x}_{c},{y}_{c}))
\end{equation}
It computes the $LBP_{P,R}$ distribution for a given $N\times M$ image.
sample.
\subsection{Color Layout Descriptor (CLD)} CLD is designed to capture the spatial distribution of color in an image. The CLD is a very compact and resolution-invariant Representation of color for high-speed image retrieval and was Designed to efficiently represent the spatial distribution of colors. The extraction procedure of CLD consist of $4$ stages: image partitioning, representative color selection, DCT transformation, and zigzag scanning. Finally we obtain the $3$ zigzag scanned matrices $(DY,DCb,DCr)$. The distance $(D)$ between two CLD descriptors is computed as:
\begin{equation}
D=\sqrt[]{\sum_{i} w_{yi}(DY_i - DY_i)^2}+\sqrt[]{\sum_{i} w_{bi}(DCb_i - DCb_i)^2}+\sqrt[]{\sum_{i} w_{ri}(DCr_i - DCr_i)^2}
\end{equation}
Where, $w_{yi},w_{bi},w_{ri}$ are the weight matrices and $DY_i, DCb_i, DCr_i$ is the $i^{th}$ element of the $3$ matrices respectively.
\subsection{Edge Oriented Histogram (EOH)} The primary objective of EOH is to first detect the edges present in an image and then construct a histogram of the directions of the gradients of those edges. First edges are detected through the Sobel operator. An image is convoluted using these masks and results in matrices indicating the edge strength of the edge for a given orientation. A histogram is created from these images by counting the maximum gradient in the five different directions.

\section{Indexing}
\label{sec:index}
\begin{figure}
\centering
\includegraphics[scale=0.7]{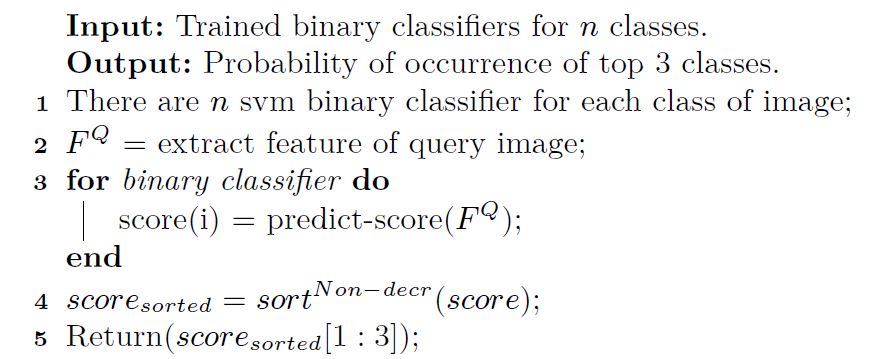}
\caption{Algorithm for class specific feature selection.}
\label{fig:algo1}
\end{figure}

Image indexing facilitates reducing the search space for similar images for a given query image. Given $n$ number of image categories, our proposed technique reduces the search space to $k$ categories, where $k < n$. We have used multiclass support vector machines (SVM) to classify the top $3$ categories having images similar to query image (see Alg. \ref{fig:algo1}). For each category we have created a binary SVM to predict whether a test image belongs to category of binary support vector machine or not. The SVM classification score for classifying observation $x$ is the signed distance from $x$ to the decision boundary ranging from $-\infty$ to $+\infty$. A positive score for a class indicates that $x$ is predicted to be in that class, a negative score indicates otherwise. We computed SVM classification scores for query image using all binary support vector machines. Then we search for similar images in reduced search space.

\section{Automatic Weight Selection}
\label{sec:weight}
For a given image importance of a feature descriptor is different. Therefore, we used a combination of feature descriptors to find the images similar to a query. We have used CDH, LBP, CLD and EOH as feature descriptors (see Sec. \ref{sec:feature} for details). The task is to assign appropriate weight to the individual descriptors. In this paper we are proposing two methods to achieve automatic weight selection. Following sub-sections give details of those two methods.

\subsection{Method 1: Relevant Ratio Technique}
\begin{figure}
\centering
\includegraphics[scale=0.6]{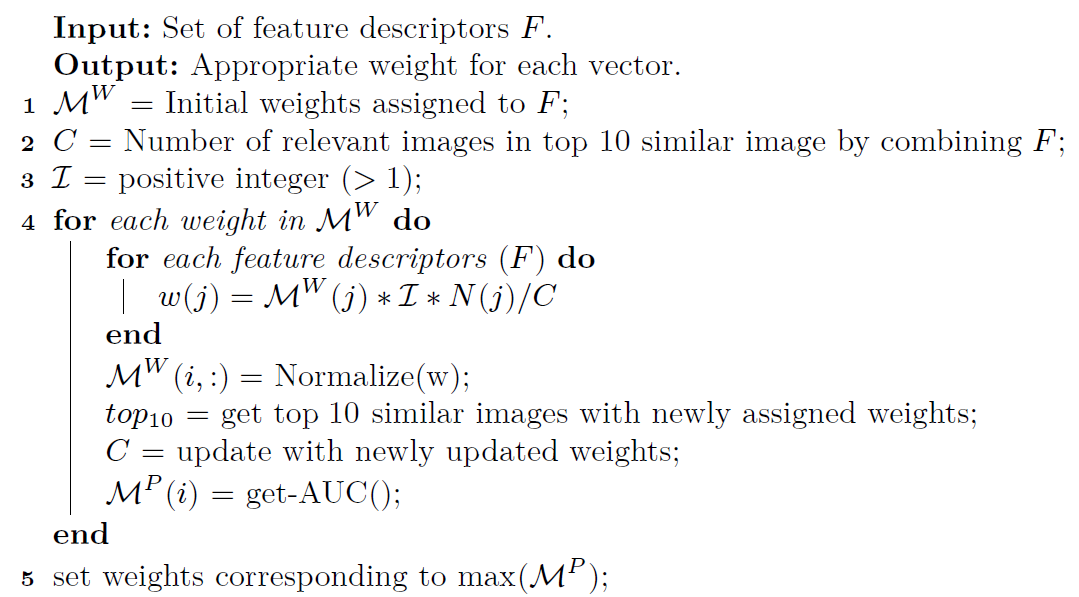}
\caption{Algorithm depicting the Relevant Ratio Method.}
\label{fig:algo2}
\end{figure}

Initial weights are assigned to the descriptors by unit normalising a set of area under the PR curve for a particular query image. Then we updated these assigned weights in each iteration as:
\begin{equation}
W_{F(i+1)} = W_{F(i)}*IF*(K_{F} / K_{C})
\end{equation}
where, $i$ is the iteration number, $K_F$ is the number of relevant images in top $10$ similar images when one feature descriptor is used only, $K_C$ is the number of relevant images in top $10$ similar images when combination of feature descriptor is used with recently updated weights, $IF$ is increment factor (any positive value greater than $1$). Algorithm \ref{fig:algo2} outlines the steps to be followed for the proposed automatic feature weight update method using the relevance ratio.

\subsection{Method 2: Mean Difference Method}
\begin{figure}[!b]
\centering
\includegraphics[scale=0.6]{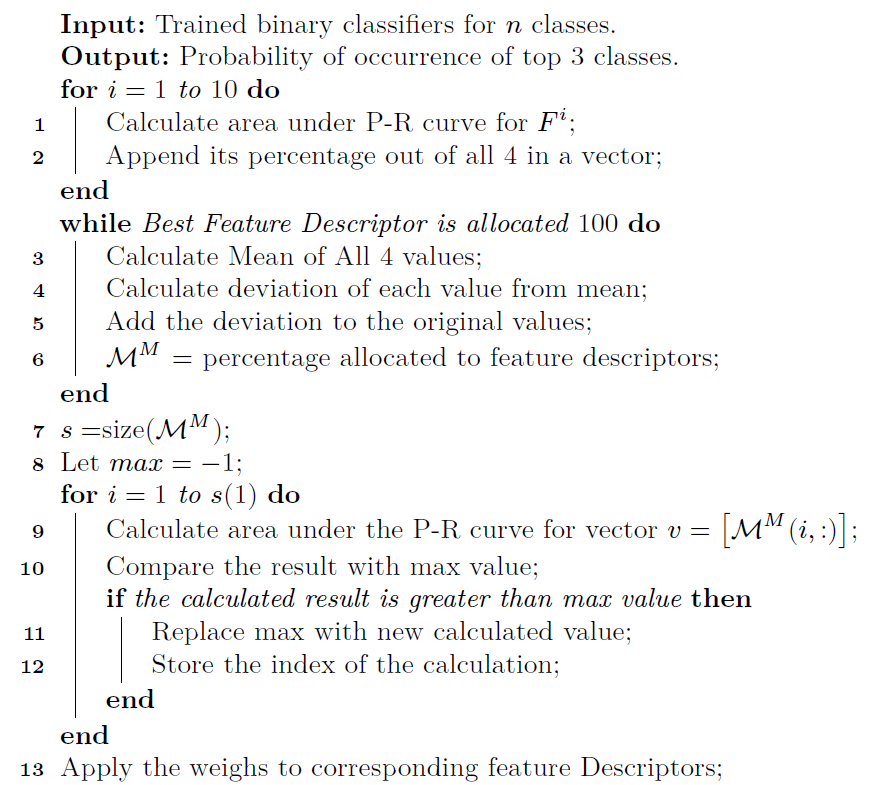}
\caption{Algorithm for automatic weight selection.}
\label{fig:algo3}
\end{figure}

Algorithm \ref{fig:algo3} gives the steps of the proposed mean difference technique. Based on the total value of all the descriptors, their initial value is allocated according to their individual weighs over the total weight. By this we increase the contribution of the feature descriptor having greater initial weight allocated to it. This technique converges till the best feature descriptor (one with maximum initial weight) is allocated $100\%$. Until that we store the weights allocated at each iteration in a matrix and finally calculate that weights which gives maximum P-R value among all iterations. 

 \begin{table}[t]
\centering
\caption{Summary of image data sets used}
\label{tab:data}
\begin{tabular}{ |p{3cm}|p{2.5cm}|p{2cm}|p{2cm}|p{1cm}|  }
\hline
Database Name& Image dimension & No. of Class &Samples/ Class &Total\\
\hline
Oxford Flower (D1)& $256\times256$ &17 &80 &1360\\ \hline
Natural Images (D2) & $256\times256$ & 16 &40 &640\\ \hline
Corel 1 (D3) & $256\times256$ & 16 &40 &640\\ \hline
Corel 2 (D4) & $256\times256$ & 16 &40 &640\\
\hline
\end{tabular}
\end{table}
\section{Results}
\label{sec:result}
\subsection{Experimental setup}
 We have used three image data sets namely, natural image data set, two subsets of Corel image data sets \cite{corel}. Each data set (see Tab. \ref{tab:data} ) is comprised of total $640$ images of $16$ different categories. Images have been resized to dimension $256 \times 256$ for feature extraction. Retrieval accuracy depends not only on strong feature representation but also on good similarity measures or distance metrics. With these distance metrics we compute content similarity of images. For each image in dataset an $M$ dimensional feature vector $h_{1i}=[h_{11},h_{12}, ...., h_{1m}]$ is extracted and stored in database. Let $ h_{2i}=[h_{21},h_{22}, ...., h_{2m}] $ be a feature vector of the query image. Table \ref{metric} lists out the different distance metric used to calculate the distance between a pair of images. Here $\textit{U}_T=\sum_{i=1}^{m} \frac{h_{1i}}{m}$ and $ \textit{U}_Q=\sum_{i=1}^{m} \frac{h_{2i}}{m}$.

\begin{table}
\centering
\caption{Distance metrics used as measure of similarity}
\begin{tabular}{|p{3cm}|p{6cm}|}
\hline
Metric & Formulae \\ \hline
Canberra (M1) & $\textit{D}(h_1,h_2)=\sum_{i=1}^{m} \frac{|h_{1i} + h_{2i} |}{|h_{1i} + \textit{U}_T | + |h_{2i} + \textit{U}_Q |} $ \\\hline
Chi-square (M2) &  $\textit{D}(h_1,h_2)=\sum_{i=1}^{m} \frac{\big(h_{1i} - h_{2i} \big)^2}{\frac{\big(h_{1i} + h_{2i}\big)}{2} } $ \\\hline
Euclidean (M3)& $\textit{D}(h_1,h_2)=\Bigg(\displaystyle\sum_{i=1}^{m} \mid h_{1i}  -  h_{2i} \mid  \Bigg)^\frac{1}{2}$\\\hline
\end{tabular}
\label{metric}
\end{table}
\subsection{Performance metrics}
Precision-Recall (PR) graph is used for measuring the accuracy of our proposed CBIR  system. Precision and Recall are based on understanding and measure of relevance. Given a set of retrieved images, the measure of precision shows the percentage of relevant samples retrieved by a search engine. On the other hand, recall signifies that out of the total number of the relevant samples for a given class, how many samples are retrieved by the system. For an ideal system for higher value of recall the system should yield higher precision. In our experiments, we have calculated the precision recall values for $100$ randomly selected query samples from the test set. After that the recall values are normalized to a scale of $0$ to $1$ and the corresponding precision values are interpolated. This is repeated for all the distance metrics used in our system.

\subsection{Qualitative Results}
The figures below shows top 8 images(similar) retrieved with respect to our Query image shown on left. Results obtained by employing different distance metrices are shown hierarchically one after the other. The images are arranged sequentially according to their ranking of similarity.Correctly retrieved images are outlined as green whereas false images are highlighted with red border. 

\begin{figure}[t]
\centering
    \includegraphics[width=\linewidth]{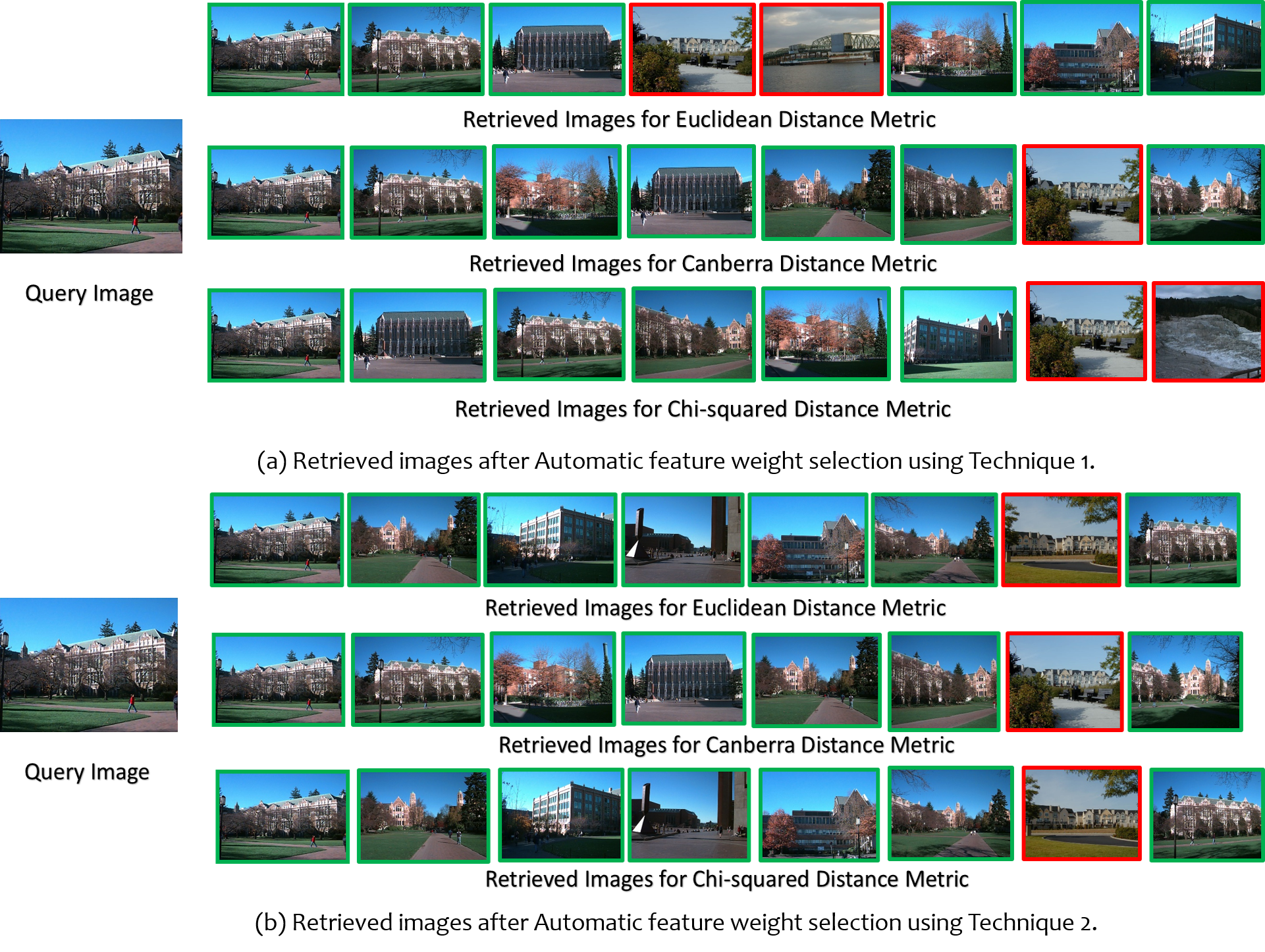}
    \caption{Retrieval Results.}
\end{figure}

\begin{table}[]
    \centering
    \begin{tabular}{|c|c|c|c|c|c|c|c|c|c|}
    \hline
         Feature Metric& \multicolumn{3}{|c|}{Dataset 1} & \multicolumn{3}{|c|}{Dataset 2} & \multicolumn{3}{|c|}{Dataset 3} \\ \hline
         CDH&36.74&36.74&36.75&0.00  &0.00 &0.00&21.1185  &21.1185 &21.1185\\\hline
         LBP&1.37&0.03&1.375&0.00  &0.00 &0.00&27.426  &27.426 &27.426\\\hline
         CLD&22.19&36.74&36.75&100.00  &100.00 &100.00&38.3047  &38.3047 &38.3047\\\hline
         EOH&39.72&36.74&36.75&0.00  &0.00 &0.00&13.1507  &13.1507 &13.1507\\\hline
         AUC&0.97&36.74&36.75&0.00  &0.00 &0.00&0.995 &0.935 &0.925\\\hline
    \end{tabular}
    \caption{Optimal weights assigned to feature descriptors for various Datasets.}
    \label{tab:weigth}
\end{table}

\subsection{Quantitative Results}
We took $50$ random images from each image dataset. We plotted average precision recall curve for each individual feature descriptors and then for combined feature descriptors. We did this using Canberra (M1), chi-square (M2) and euclidean distance (M3) metrics. Table \ref{auc1} shows the value of area under the curve of precision recall curve for three datasets that we used. Value of area under the precision recall curve is the measure of accuracy of retrieval systems. Highest accuracy was achieved when we combined feature descriptors and assigned optimal weights using relevant ratio technique.

\begin{table*}
\centering
\caption{Area under the P-R curve for various datasets using all possible distance metric and feature combination.}
\label{auc1}
\begin{tabular}{|l|l|l|l|l|l|l|}
\hline
\multicolumn{1}{|l|}{\textbf{Data}} & \textbf{Metric} & \textbf{CDH} & \textbf{LBP} & \textbf{CLD} & \textbf{EOH} & \textbf{Comb.} \\ \hline
\multirow{3}{*}{D1}         & M1                & 0.854      & 0.728      & 0.618 & 0.655      & \textbf{0.941}  \\ \cline{2-7} 
                                       & M2               & 0.846      & 0.717      & 0.610      & 0.609      & \textbf{0.928}  \\ \cline{2-7} 
                                       & M3                & 0.823       & 0.723      & 0.599 & 0.617      & \textbf{0.905}  \\ \hline
\multirow{3}{*}{D2}         & M1                & 0.854      & 0.728      & 0.618 & 0.655      & \textbf{0.941}  \\ \cline{2-7} 
                                       & M2               & 0.846      & 0.717      & 0.610      & 0.609      & \textbf{0.928}  \\ \cline{2-7} 
                                       & M3                & 0.823       & 0.723      & 0.599 & 0.617      & \textbf{0.905}  \\ \hline
\multirow{3}{*}{D3}        & M1                 & 0.817      & 0.739      & 0.703 & 0.669      & \textbf{0.951}  \\ \cline{2-7} 
                                       & M2               & 0.793      & 0.724      & 0.704 & 0.619      & \textbf{0.919}   \\ \cline{2-7} 
                                       & M3                & 0.801      & 0.715      & 0.689      & 0.595      & \textbf{0.909}  \\ \hline
\multirow{3}{*}{D4}        & M1                & 0.787      & 0.727      & 0.671 & 0.683     & \textbf{0.895}  \\ \cline{2-7} 
                                       & M2               & 0.753      & 0.711      & 0.673      & 0.635      & \textbf{0.879}  \\ \cline{2-7} 
                                       & M3                & 0.705        & 0.724      & 0.686 & 0.623      & \textbf{0.878}  \\ \hline
\end{tabular}
\end{table*}

\begin{figure}
    \centering
    \includegraphics[width=\linewidth]{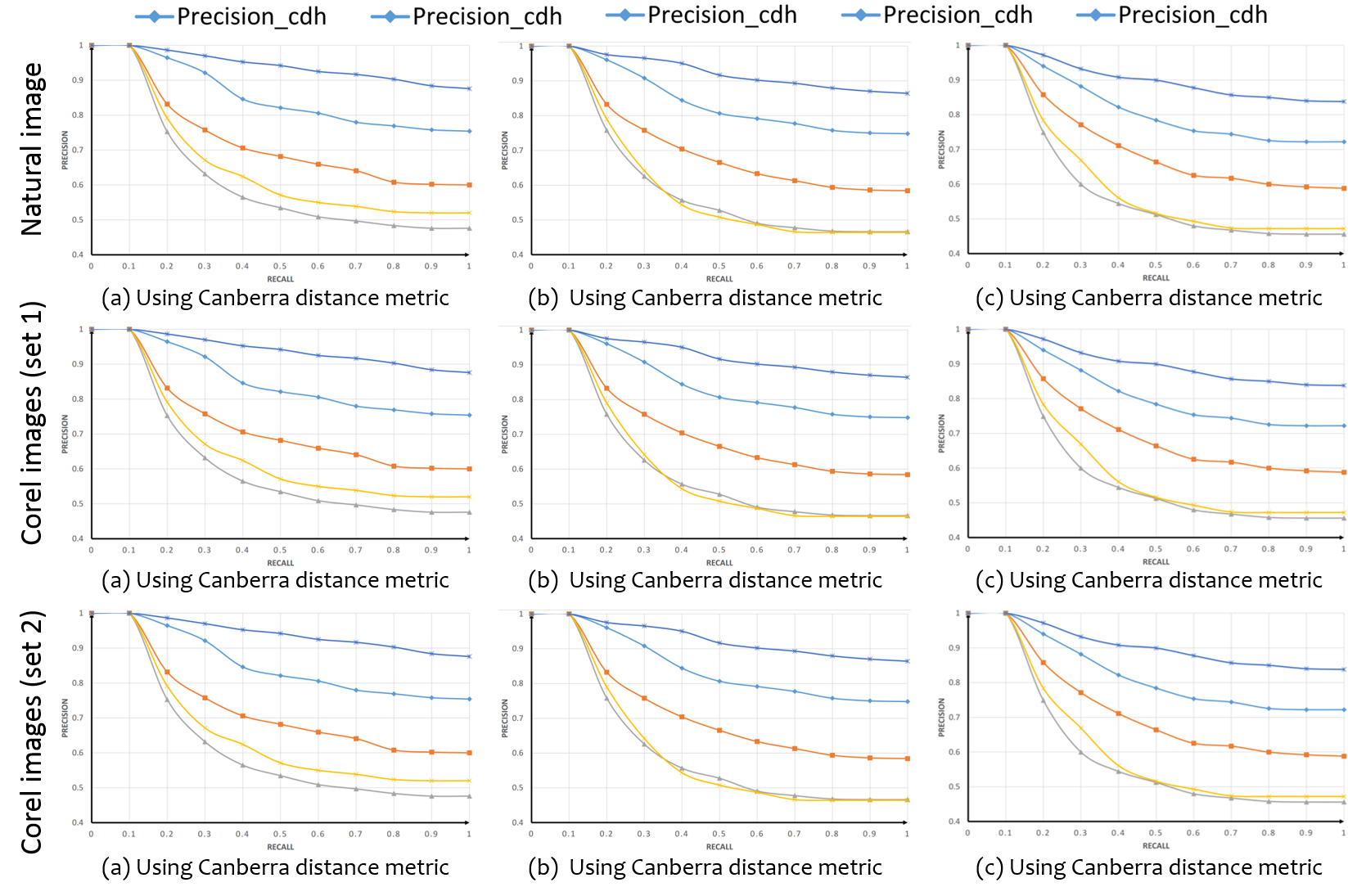}
    \caption{Precision-Recall plot depicting the quantitative performance of the proposed method.}
    \label{fig:pr_image}
\end{figure}

For each of the dataset average precision recall curve for individual feature descriptors and combined feature descriptors using the three different distance metrics are shown in Fig. \ref{fig:pr_image}. On horizontal axis we have recall and vertical axis we have precision. Individually color difference histogram is the best feature descriptor. Curve corresponding to combined feature descriptors have highest value of area under the precision curve in each plot.

\section{Discussion}
\label{sec:discuss}
We have implemented image-indexing and automated weight selection. Using image indexing we are predicting top 3 categories of images that have images similar to query image. By doing so we are reducing our search space by 81.25\%.  This prediction have impact while we search for top 10 similar images. In automated weight selection, we are initially assigning weights to feature descriptors by unit normalizing the values of area under the precision recall curves for corresponding feature descriptors. We have used two techniques to automate these weights.

With automatically assigned weights to feature descriptors, we are getting much accurate results than individual feature descriptors. We can see in precision recall curves(above) the area under the precision recall curve is more for automatically assigned weights than individual feature descriptors. We are able to find the most important feature descriptor for a particular query image out of four feature descriptors that we have used. We can see this in tables in Qualitative results section. Most important feature descriptors have been assigned the maximum weights.

\section{Conclusion}
\label{sec:concl}
With the advent of various search engines, image searching has become an easier task. But search engines mostly use text based retrieval techniques. Though CBIR is a happening topic, we cannot expect the entire upheaval of existing techniques with CBIR. But certainly, CBIR can be used to complement the existing machinery to provide better results. The CBIR method presented herein use combination of local and global features . The purpose of this paper was to improve the accuracy of a CBIR application by allowing the system to retrieve more images similar to the query image. The proposed methodology first use image indexing to reduce the search space for similar images to query image, then assign optimal weights to individual feature to use feature combination. It also predict the most important feature( out of used ones ) to a query image.




\end{document}